\documentclass[11pt,letterpaper]{article}
\usepackage{emnlp2016}
\usepackage{times}
\usepackage{latexsym}
\usepackage{amsmath}
\usepackage{amssymb}
\usepackage{graphicx}
\usepackage{xcolor}
\usepackage{array}

\emnlpfinalcopy

\def\emnlppaperid{889}

\usepackage{microtype}
\usepackage{booktabs}

\usepackage{hyperref}

\newcommand{\ab}{\mathbf{a}}
\newcommand{\bb}{\mathbf{b}}
\newcommand{\vb}{\mathbf{v}}
\newcommand{\yhat}{\hat{y}}
\newcommand{\yb}{\mathbf{y}}
\newcommand{\thetab}{\mathbf{\theta}}
\newcommand{\RR}{\mathbb{R}}
\newcommand{\argmax}{\textrm{argmax}}

\newcommand{\eat}[1]{\ignorespaces}

\newcolumntype{L}{>{\arraybackslash}m{7cm}}

\title{A Decomposable Attention Model for Natural Language Inference}

\author{Ankur P. Parikh \\ Google \\ New York, NY   \And
Oscar T{\"{a}}ckstr{\"{o}}m \\ Google \\ New York, NY   \And
Dipanjan Das \\ Google \\ New York, NY   \And
Jakob Uszkoreit \\ Google \\ Mountain View, CA  \ANDEMAIL { {\tt \{aparikh,oscart,dipanjand,uszkoreit\}@google.com} } }

\date{}

\begin{document}

\maketitle

\begin{abstract}
We propose a simple neural architecture for natural language inference.
Our approach uses attention to decompose the problem into subproblems that can be solved separately, thus making it trivially parallelizable.
On the Stanford Natural Language Inference (SNLI) dataset, we obtain state-of-the-art results with almost an order of magnitude fewer parameters than previous work and without relying on any word-order information.
Adding intra-sentence attention that takes a minimum amount of order into account yields further improvements.
\end{abstract}

\vspace{-0.1cm}
\section{Introduction}
\vspace{-0.1cm}

Natural language inference (NLI) refers to the problem of determining entailment and contradiction relationships between a premise and a hypothesis.
NLI is a central problem in language understanding \cite{katz1972semantic,bosmarkert2005,benthem2008brief,MacCartney:2009} and
recently the large SNLI corpus of 570K sentence pairs was created for this task \cite{bowman2015large}.
We present a new model for NLI and leverage this corpus for comparison with prior work.

A large body of work based on neural networks for text similarity tasks including NLI has been published in recent years \cite[\textit{inter alia}]{hu2014convolutional,rocktaschel2015reasoning,wang2015learning,yin2015abcnn}.
The dominating trend in these models is to build complex, deep text representation models, for example, with convolutional networks \cite[CNNs henceforth]{le1990handwritten} or long short-term memory networks \cite[LSTMs henceforth]{hochreiter1997long} with the goal of deeper sentence comprehension. While these approaches have yielded impressive results, they are often computationally very expensive, and result in models having millions of parameters (excluding embeddings).

Here, we take a different approach, arguing that for natural language inference it can often suffice to simply align bits of local text substructure and then aggregate this 
information. For example, consider the following sentences:

\begin{itemize}
\vspace{-0.05cm}
\it
\setlength\itemsep{-0.05em}
\item Bob is in his room, but because of the thunder and lightning outside, he cannot sleep.
\item Bob is awake.
\item It is sunny outside.
\vspace{-0.05cm}
\end{itemize}

The first sentence is complex in structure and it is challenging to construct a compact representation that expresses its entire meaning.
However, it is fairly easy to conclude that the second sentence follows from the first one, by simply aligning \textit{Bob} with \textit{Bob} and \textit{cannot sleep} with \textit{awake} and recognizing that these are synonyms.
Similarly, one can conclude that \textit{It is sunny outside} contradicts the first sentence, by aligning \textit{thunder and lightning} with \textit{sunny} and recognizing that these are most likely incompatible.

We leverage this intuition to build a simpler and more lightweight approach to NLI within a neural framework; with considerably fewer parameters, our model outperforms more complex existing neural architectures.
In contrast to existing approaches, our approach only relies on alignment and is fully computationally decomposable with respect to the input text.
An overview of our approach is given in Figure~\ref{fig:overview}.
Given two sentences, where each word is represented by an embedding vector, we first create a soft alignment matrix using neural attention \cite{bahdanau2014neural}.
We then use the (soft) alignment to decompose the task into subproblems that are solved separately.
Finally, the results of these subproblems are merged to produce the final classification.
In addition, we optionally apply intra-sentence attention \cite{cheng2016long} to endow the model with a richer encoding of substructures prior to the alignment step.

Asymptotically our approach does the same total work as a vanilla LSTM encoder, while being trivially parallelizable across sentence length, which can allow for considerable speedups in low-latency settings.
Empirical results on the SNLI corpus show that our approach achieves state-of-the-art results, while using almost an order of magnitude fewer parameters compared to complex LSTM-based approaches.

\vspace{-0.1cm}
\section{Related Work}
\vspace{-0.1cm}

Our method is motivated by the central role played by alignment in machine translation \cite{koehn2009statistical} and previous approaches to sentence similarity modeling \cite{haghighi2005robust,das2009paraphrase,chang2010discriminative,fader2013paraphrase}, natural language inference~\cite{marsi2005classification,maccartney2006learning,hickl2007discourse,maccartney2008phrase}, and semantic parsing~\cite{andreas2013semantic}.
The neural counterpart to alignment, \emph{attention} \cite{bahdanau2014neural}, which is a key part of our approach, was originally proposed and has been predominantly used in conjunction with LSTMs \cite{rocktaschel2015reasoning,wang2015learning} and to a lesser extent with CNNs \cite{yin2015abcnn}.
In contrast, our use of attention is purely based on word embeddings and our method essentially consists of feed-forward networks that operate largely independently of word order.

\vspace{-0.1cm}
\section{Approach}
\label{sec:approach}
\vspace{-0.1cm}

Let $\ab = (a_1,\ldots,a_{\ell_a})$ and $\bb = (b_1,\ldots,b_{\ell_b})$ be the two input sentences of length $\ell_a$ and $\ell_b$, respectively.
We assume that each $a_i$, $b_j \in \RR^d$ is a word embedding vector of dimension $d$ and that each sentence is prepended with a ``NULL" token. Our training data comes in the form of labeled pairs $\{\ab^{(n)}, \bb^{(n)}, \yb^{(n)} \}_{n=1}^N$, where $\yb^{(n)} = (y_1^{(n)},\ldots,y_C^{(n)})$ is an indicator vector encoding the label and $C$ is the number of output classes.
At test time, we receive a pair of sentences $(\ab, \bb)$ and our goal is to predict the correct label $\yb$.

\paragraph{Input representation.} Let $\bar{\ab} = (\bar{a}_1,\ldots,\bar{a}_{\ell_a})$ and $\bar{\bb} = (\bar{b}_1,\ldots, \bar{b}_{\ell_b})$ denote the input representation of each fragment that is fed to subsequent steps of the algorithm. The vanilla version of our model simply defines $\bar{\ab} := \ab$ and $\bar{\bb} := \bb$. With this input representation, our model does not make use of word order.
However, we discuss an extension using intra-sentence attention in Section~\ref{subsec:intra} that uses a minimal amount of sequence information.

The core model consists of the following three components (see Figure~\ref{fig:overview}), which are trained jointly:

\begin{figure}[!tbp]
\includegraphics[width=\columnwidth]{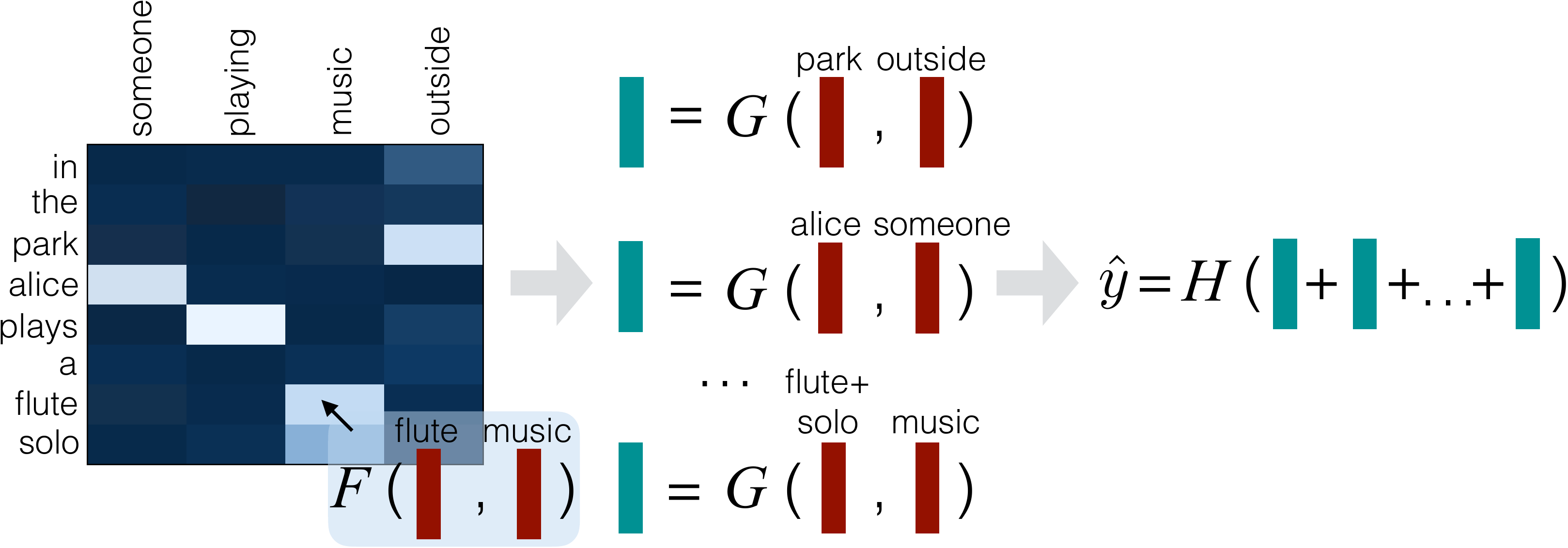}
\caption{Pictoral overview of the approach, showing the \emph{Attend} (left), \emph{Compare} (center) and \emph{Aggregate} (right) steps.}
\label{fig:overview}
\vspace{-0.3cm}
\end{figure}

\paragraph{Attend.} First, soft-align the elements of $\bar{\ab}$ and $\bar{\bb}$ using a variant of neural attention \cite{bahdanau2014neural} and decompose the problem into the comparison of aligned subphrases.

\paragraph{Compare.} Second, separately compare each aligned subphrase to produce a set of vectors $\{\vb_{1,i}\}_{i=1}^{\ell_a}$ for $\ab$ and
$\{\vb_{2,j}\}_{j=1}^{\ell_b}$ for $\bb$.
Each $\vb_{1,i}$ is a nonlinear combination of $a_i$ and its (softly) aligned subphrase in $\bb$
(and analogously for $\vb_{2,j}$).

\paragraph{Aggregate.} Finally, aggregate the sets $\{\vb_{1,i}\}_{i=1}^{\ell_a}$ and $\{\vb_{2,j}\}_{j=1}^{\ell_b}$ from the previous step and use the result to predict the label $\yhat$.

\subsection{Attend}
We first obtain unnormalized attention weights $e_{ij}$, computed by a function $F'$, which decomposes as:
\begin{align}\label{eq:unnormalized_attention}
e_{ij} := F'(\bar{a}_i, \bar{b}_j) := F(\bar{a}_i)^{T} F(\bar{b}_j)\,.
\end{align}
This decomposition avoids the quadratic complexity that would be associated with
separately applying $F'$ $\ell_a \times \ell_b$ times. Instead, only $\ell_a + \ell_b$ applications of $F$ are needed.
We take $F$ to be a feed-forward neural network with ReLU activations \cite{glorot2011deep}.

These attention weights are normalized as follows:
\begin{align}\label{eq:normalized_attention}
\beta_i &:= \sum_{j=1}^{\ell_b} \frac{\exp(e_{ij})}{\sum_{k=1}^{\ell_b} \exp(e_{ik})} \bar{b}_j\,,  \notag \\
\alpha_j &:= \sum_{i=1}^{\ell_a} \frac{\exp(e_{ij})}{\sum_{k=1}^{\ell_a} \exp(e_{kj})} \bar{a}_i\,.
\end{align}
Here $\beta_i$ is the subphrase in $\bar{\bb}$ that is (softly) aligned to $\bar{a}_i$ and vice versa for $\alpha_j$.

\subsection{Compare}
Next, we separately compare the aligned phrases $\{(\bar{a}_i, \beta_i)\}_{i=1}^{\ell_a}$ and $\{(\bar{b}_j, \alpha_j)\}_{j=1}^{\ell_b}$ using a function $G$, which in this work is again a feed-forward network:
\begin{align}
\vb_{1,i} &:= G([\bar{a}_i, \beta_i])\quad \forall i \in [1,\ldots, \ell_a]\,, \notag \\
\vb_{2,j} &:= G([\bar{b}_j, \alpha_j])\quad \forall j \in [1,\ldots, \ell_b]\,.
\label{eq:gdefine}
\end{align}
where the brackets $[\cdot, \cdot]$ denote concatenation. Note that since there are only a linear number of terms in this case, we do not need to apply a decomposition as was done in the previous step. Thus $G$ can jointly take into account both $\bar{a}_i,$ and $\beta_i$.

\subsection{Aggregate}
We now have two sets of comparison vectors $\{\vb_{1,i}\}_{i=1}^{\ell_a}$ and $\{\vb_{2,j}\}_{j=1}^{\ell_b}$. We first aggregate over each set by summation:
\begin{align}
\vspace{-0.1cm}
\vb_{1} &= \sum_{i=1}^{\ell_a} \vb_{1,i} \qquad\,, \qquad \vb_{2} = \sum_{j=1}^{\ell_b}  \vb_{2,j}\,.
\vspace{-0.1cm}
\end{align}
and feed the result through a final classifier $H$, that is a feed forward network followed by a linear layer:
\begin{align}
\hat{\yb} = H([\vb_1, \vb_2])\,,
\label{eq:hdefine}
\end{align}
where $\hat{\yb} \in \RR^C$ represents the predicted (unnormalized) scores for each class and consequently the predicted class is given by $\yhat = \argmax_i \hat{\yb}_i$.

For training, we use multi-class cross-entropy loss with dropout regularization~\cite{srivastava2014dropout}:
\begin{align}
\notag
L(\thetab_F, \thetab_G, \thetab_H) = \frac{1}{N} \sum_{n=1}^{N} \sum_{c=1}^{C} y_c^{(n)} \log \frac{\exp(\yhat_c)}{\sum_{c'=1}^C \exp(\yhat_{c'})}\,.
\end{align}
Here $\thetab_F, \thetab_G, \thetab_H$ denote the learnable parameters of the functions F, G and H, respectively.


\subsection{Intra-Sentence Attention (Optional)}
\label{subsec:intra}
In the above model, the input representations are simple word embeddings. 
However, we can augment this input representation with \emph{intra-sentence attention} to encode compositional relationships between words within each sentence, as proposed by \newcite{cheng2016long}.
Similar to Eqs.~\ref{eq:unnormalized_attention} and~\ref{eq:normalized_attention}, we define
\begin{align}
f_{ij} := F_{\textrm{intra}}(a_i)^{T} F_{\textrm{intra}}(a_j)\,,
\end{align}
where $F_{\textrm{intra}}$ is a feed-forward network. We then create the self-aligned phrases
\begin{align}
a'_i &:= \sum_{j=1}^{\ell_a} \frac{\exp(f_{ij} + d_{i-j})}{\sum_{k=1}^{\ell_a} \exp(f_{ik} + d_{i-k})} a_j\,.
\end{align}
The distance-sensitive bias terms $d_{i-j}\in \RR$ provides the model with a minimal amount of sequence information, while remaining parallelizable.
These terms are bucketed such that all distances greater than 10 words share the same bias.
The input representation for subsequent steps is then defined as $\bar{a}_i := [a_i, a'_i]$ and analogously $\bar{b}_i := [b_i, b'_i]$.

\begin{table*}[!tbp]
\begin{center}
\begin{tabular}{lccr}
\toprule
Method & Train Acc & Test Acc & \#Parameters \\
\midrule
Lexicalized Classifier \cite{bowman2015large} & 99.7 & 78.2 & --  \\
\midrule
300D LSTM RNN encoders \cite{bowman2016fast} & 83.9 & 80.6 & 3.0M \\
1024D pretrained GRU encoders \cite{vendrov2015order} & 98.8 & 81.4 & 15.0M \\
300D tree-based CNN encoders \cite{mou2015recognizing} & 83.3 & 82.1 & 3.5M \\
300D SPINN-PI encoders \cite{bowman2016fast} & 89.2 & 83.2 &  3.7M \\
\midrule
100D LSTM with attention \cite{rocktaschel2015reasoning} & 85.3 & 83.5 & 252K \\
300D mLSTM \cite{wang2015learning} & 92.0 & 86.1 & 1.9M \\
450D LSTMN with deep attention fusion \cite{cheng2016long} & 88.5 & 86.3 & 3.4M \\
\midrule
Our approach (vanilla)  & 89.5 & 86.3 & 382K \\
Our approach with intra-sentence attention & 90.5 & \textbf{86.8} & 582K \\
\bottomrule
\end{tabular}
\end{center}
\caption{Train/test accuracies on the SNLI dataset and number of parameters (excluding embeddings) for each approach.}
\label{table:snli-results}
\vspace{-0.4cm}
\end{table*}

\vspace{-0.1cm}
\section{Computational Complexity}
\label{sec:complexity}
\vspace{-0.1cm}

We now discuss the asymptotic complexity of our approach and how it offers a higher degree of parallelism than LSTM-based approaches.
Recall that $d$ denotes embedding dimension and $\ell$ means sentence length. For simplicity we assume that all hidden dimensions are $d$ and that the complexity of matrix($d\times d$)-vector($d \times 1$) multiplication is $O(d^2)$.

A key assumption of our analysis is that $\ell < d$, which we believe is reasonable and is true of the SNLI dataset \cite{bowman2015large} where $\ell < 80$, whereas recent LSTM-based approaches have used $d \ge 300$. This assumption allows us to bound the complexity of computing the $\ell^2$ attention weights.

\paragraph{Complexity of LSTMs.} The complexity of an LSTM cell is $O(d^2)$, resulting in a complexity of $O(\ell d^2)$ to encode the sentence. Adding attention as in \newcite{rocktaschel2015reasoning} increases this complexity to $O(\ell d^2 + \ell^2 d)$.

\paragraph{Complexity of our Approach.} Application of a feed-forward network requires $O(d^2)$ steps.
Thus, the \textbf{Compare} and \textbf{Aggregate} steps have complexity $O(\ell d^2)$ and $O(d^2)$ respectively. For the \textbf{Attend} step, $F$ is evaluated $O(\ell)$ times, giving a complexity of $O(\ell d^2)$. Each attention weight $e_{ij}$ requires one dot product, resulting in a complexity of $O(\ell^2 d)$.

Thus the total complexity of the model is $O(\ell d^2 + \ell^2 d)$, which is equal to that of an LSTM with attention.
However, note that with the assumption that $\ell < d$, this becomes $O(\ell d^2)$ which is the same complexity as a regular LSTM. Moreover, unlike the LSTM, our approach has the advantage of being parallelizable over $\ell$, which can be useful at test time.

\vspace{-0.1cm}
\section{Experiments}
\label{sec:experiments}
\vspace{-0.1cm}

\begin{table}
\begin{tabular}{lccc}
\toprule
Method & \textbf{N}  & \textbf{E}  & \textbf{C} \\
\midrule
\newcite{bowman2016fast} & 80.6 & 88.2 & 85.5 \\
\newcite{wang2015learning} & 81.6 & 91.6 & 87.4 \\
Our approach (vanilla) & 83.6 & 91.3 & 85.8 \\
Our approach w/ intra att. & 83.7 & 92.1 & 86.7 \\
\bottomrule
\end{tabular}
\caption{Breakdown of accuracy with respect to classes on SNLI development set. 
\textbf{N}=neutral,  \textbf{E}=entailment, \textbf{C}=contradiction.}
\label{table:all-errors}
\vspace{-0.4cm}
\end{table}

\begin{table*}
\begin{center}
\resizebox{\textwidth}{!}{
\begin{tabular}{cLLccccc}
\toprule
ID & Sentence 1 & Sentence 2 & DA (vanilla) & DA (intra att.) &  SPINN-PI  & mLSTM & Gold  \\
\midrule
A & Two kids are standing in the ocean hugging each other. &  Two kids enjoy their day at the beach. & \textbf{N} & \textbf{N} & E & E & \textbf{N} \\
B & A dancer in costumer performs on stage while a man watches. & the man is captivated & \textbf{N} & \textbf{N} & E & E & \textbf{N} \\
C & They are sitting on the edge of a fountain & The fountain is splashing the persons seated. & \textbf{N} & \textbf{N} & C & C & \textbf{N} \\
\midrule
D & Two dogs play with tennis ball in field. & Dogs are watching a tennis match. & N & \textbf{C} & \textbf{C} & \textbf{C} & \textbf{C} \\
E & Two kids begin to make a snowman on a sunny winter day. &  Two penguins making a snowman. & N & \textbf{C} & \textbf{C} & \textbf{C} & \textbf{C} \\
F & The horses pull the carriage, holding people and a dog, through the rain. & Horses ride in a carriage pulled by a dog. & E & E & \textbf{C} & \textbf{C} & \textbf{C} \\
\midrule
G & A woman closes her eyes as she plays her cello. &  The woman has her eyes open. & E & E & E & E & \textbf{C} \\
H & Two women having drinks and smoking cigarettes at the bar. & Three women are at a bar. & E & E & E & E & \textbf{C} \\
I & A band playing with fans watching. & A band watches the fans play & E & E & E & E & \textbf{C} \\
\bottomrule
\end{tabular}}
\end{center}
\caption{Example wins and losses compared to other approaches. DA (Decomposable Attention) refers to our approach while SPINN-PI and mLSTM are previously developed methods (see Table 1).}
\label{table:error-analysis}
\vspace{-0.4cm}
\end{table*}

We evaluate our approach on the Stanford Natural Language Inference (SNLI) dataset \cite{bowman2015large}.
Given a sentences pair $(\ab, \bb)$, the task is to predict whether $\bb$ is \textit{entailed} by $\ab$, $\bb$ \textit{contradicts} $\ab$, or whether their relationship is \textit{neutral}.

\subsection{Implementation Details}
The method was implemented in TensorFlow \cite{abaditensorflow}.

\textbf{Data preprocessing:} Following \newcite{bowman2015large}, we remove examples labeled ``--'' (no gold label) from the dataset, which leaves 549,367 pairs for training, 9,842 for development, and 9,824 for testing. We use the tokenized sentences from the non-binary parse provided in the dataset and prepend each sentence with a ``NULL" token. During training, each sentence was padded up to the maximum length of the batch for efficient training (the padding was explicitly masked out so as not to affect the objective/gradients). For efficient batching in TensorFlow, we semi-sorted the training data to first contain examples where both sentences had length less than 20, followed by those with length less than 50, and then the rest. This ensured that most training batches contained examples of similar length.

\textbf{Embeddings: } We use 300 dimensional GloVe embeddings \cite{pennington2014glove} to represent words. Each embedding vector was normalized to have $\ell_2$ norm of 1 and projected down to 200 dimensions, a number determined via hyperparameter tuning. Out-of-vocabulary (OOV) words are hashed to one of 100 random embeddings each initialized to mean 0 and standard deviation 1. All embeddings remain fixed during training, but the projection matrix is trained. All other parameter weights (hidden layers etc.) were initialized from random Gaussians with mean 0 and standard deviation 0.01.

Each hyperparameter setting was run on a single machine with 10 asynchronous gradient-update threads, using Adagrad \cite{duchi2011adaptive} for optimization with the default initial accumulator value of 0.1. Dropout regularization~\cite{srivastava2014dropout} was used for all ReLU layers, but not for the final linear layer. We additionally tuned the following hyperparameters and present their chosen values in parentheses: network size (2-layers, each with 200 neurons), batch size (4), \footnote{16 or 32 also work well and are a bit more stable.} dropout ratio (0.2) and learning rate (0.05--vanilla, 0.025--intra-attention). All settings were run for 50 million steps (each step indicates one batch) but model parameters were saved frequently as training progressed and we chose the model that did best on the development set.

\subsection{Results}
Results in terms of 3-class accuracy are shown in Table~\ref{table:snli-results}.
Our vanilla approach achieves state-of-the-art results with almost an order of magnitude fewer parameters than the LSTMN of~\newcite{cheng2016long}. Adding intra-sentence attention gives a considerable improvement of 0.5 percentage points over the existing state of the art. Table~\ref{table:all-errors} gives a breakdown of accuracy on the development set showing that most of our gains stem from \textit{neutral}, while most losses come from \textit{contradiction} pairs.


Table~\ref{table:error-analysis} shows some wins and losses. Examples A-C are cases where both variants of our approach are correct while both SPINN-PI~\cite{bowman2016fast} and the mLSTM~\cite{wang2015learning} are incorrect. In the first two cases, both sentences contain phrases that are either identical or highly lexically related (e.g. ``Two kids" and ``ocean / beach") and our approach correctly favors neutral in these cases. In Example C, it is possible that relying on word-order may confuse SPINN-PI and the mLSTM due to how ``fountain" is the object of a preposition in the first sentence but the subject of the second.

The second set of examples (D-F) are cases where our vanilla approach is incorrect but mLSTM and SPINN-PI are correct. Example F requires sequential information and neither variant of our approach can predict the correct class. Examples D-E are interesting however, since they don't require word order information, yet intra-attention seems to help. We suspect this may be because the word embeddings are not fine-grained enough for the algorithm to conclude that ``play/watch" is a contradiction, but intra-attention, by adding an extra layer of composition/nonlinearity to incorporate context, compensates for this.

Finally, Examples G-I are cases that all methods get wrong. The first is actually representative of many examples in this category where there is one critical word that separates the two sentences (close vs open in this case) and goes unnoticed by the algorithms. Examples H requires inference about numbers and Example I needs sequence information.

\vspace{-0.1cm}
\section{Conclusion}
\vspace{-0.1cm}

We presented a simple attention-based approach to natural language inference that is trivially parallelizable.
The approach outperforms considerably more complex neural methods aiming for text understanding.
Our results suggest that, at least for this task, pairwise comparisons are relatively more important than global sentence-level representations.

\section*{Acknowledgements}
We thank Slav Petrov, Tom Kwiatkowski, Yoon Kim, Erick Fonseca, Mark Neumann for useful discussion and Sam Bowman and Shuohang Wang for providing us their model outputs for error analysis.
\bibliography{emnlp2016}
\bibliographystyle{emnlp2016}

\end{document}